\documentclass[submission,copyright,creativecommons]{eptcs}
\providecommand{\event}{ICLP 2026} 

\usepackage{iftex}

\ifpdf
  \usepackage{underscore}         
  \usepackage[T1]{fontenc}        
\else
  \usepackage{breakurl}           
\fi

\usepackage{soul}
\usepackage{amssymb}
\usepackage{amsmath}
\usepackage{booktabs}
\usepackage{enumitem}
\usepackage{multirow}
\usepackage{graphicx}
\usepackage{color}

\usepackage{mathtools}
\usepackage{algorithm}
\usepackage{listings}
\usepackage{tikz}
\usetikzlibrary{arrows,shapes,automata,petri,positioning,calc}
\usepackage{bm}

\usepackage{pifont}
\newtheorem{example}{Example}

\newcommand*{\inlineimg}[1]{%
    \raisebox{-0.2\baselineskip}{%
        \includegraphics[
        height=0.9\baselineskip,
        width=0.9\baselineskip,
        keepaspectratio,
        ]{#1}%
    }%
}

\newcommand*{\inlineimgrot}[1]{%
    \raisebox{-0.2\baselineskip}{%
        \includegraphics[
            angle=180,
            height=0.9\baselineskip,
            keepaspectratio,
            origin=c
        ]{#1}%
    }%
}

\title{Differentiable Logic Programming to Mitigate Reasoning Shortcuts in Neurosymbolic Systems}
\author{Akihiro Takemura
\institute{National Institute of Informatics, Tokyo, Japan}
\email{atakemura@nii.ac.jp}
\and
Katsumi Inoue
\institute{National Institute of Informatics, Tokyo, Japan}
\email{inoue@nii.ac.jp}
}
\def\titlerunning{Differentiable Logic Programming to Mitigate Reasoning Shortcuts in Neurosymbolic Systems}
\def\authorrunning{A. Takemura \& K. Inoue}
\begin{document}
\maketitle

\begin{abstract}
Neurosymbolic (NeSy) systems integrate neural networks with logical reasoning to achieve both generalization and interpretability, but recent work has shown they are susceptible to shortcut reasoning behaviors. 
We propose a novel method using matrix-based differentiable logic programming to mitigate reasoning shortcuts in two phenomena: \textit{constraint satisfaction shortcuts}, where constraints are satisfied without achieving the intended task, and \textit{cognition shortcuts}, where biased data leads to semantically incorrect concept mappings despite logically sound inference.
Building on recent matrix-based logic programming semantics, we introduce design elements to mitigate shortcuts, including a unified encoding of rules and constraints in a single matrix. 
We also identify connections to fuzzy logic t-norms and empirically compare their gradient flow properties.
Through carefully designed experiments on MNIST variants, we show that one-to-one grounding of neural outputs to logical atoms significantly reduces both shortcut types compared to previous methods that rely on soft probability distributions. 
We then confirm that architectural choices in coupling symbolic knowledge with neural learning play a critical role in shortcut mitigation.
\end{abstract}

\section{Introduction}\label{sec:intro}

Neurosymbolic (NeSy) AI aims to integrate the pattern recognition capabilities of neural networks with the structured reasoning power of symbolic logic \cite{hitzlerNeuroSymbolicArtificialIntelligence2022}. 
This integration is crucial for enabling AI systems to operate in domains requiring both perceptual understanding and high-level reasoning, such as medical diagnosis and autonomous driving. 
NeSy systems promise both generalization and interpretability by combining neural learning with logical inference.

From a logic programming perspective, the key question is how the semantics of a logic program are preserved during gradient-based optimization. 
When neural outputs serve as truth value assignments over ground atoms, the fidelity of this correspondence determines whether the system learns the intended interpretation or exploits artifacts of the optimization.

However, recent studies have shown that NeSy systems are prone to shortcut reasoning\footnote{In the NeSy literature, \textit{shortcut reasoning} refers to the phenomenon where a NeSy system finds interpretations that satisfy all program rules but do not correspond to the intended semantics, which is analogous to finding unintended models of an underspecified logic program.}
, where models achieve high training accuracy by exploiting spurious patterns or overly simplistic reasoning paths rather than learning the intended concepts \cite{marconatoNotAllNeuroSymbolic2023,liLearningLogicalConstraints2022,wangLearningLatentModels2023,marconatoBEARSMakeNeuroSymbolic2024a}.
Such shortcuts undermine model robustness, especially under weak supervision or underspecified constraints.
While these phenomena have been identified in prior work, empirical analysis of how NeSy architectures handle shortcuts remains limited.

In this paper, we examine two shortcut phenomena.
\textit{Constraint satisfaction shortcuts} \cite{liLearningLogicalConstraints2022}: These occur when a model trivially satisfies logical constraints (e.g., by falsifying the antecedent of an implication to satisfy the constraint) without learning the intended concept the constraint is meant to enforce.
\textit{Cognition shortcuts} \cite{marconatoNotAllNeuroSymbolic2023}: These involve conceptual confusion, where a model trained on biased or confounded data learns incorrect mappings between inputs and concepts, leading to reasoning errors even if logical rules are syntactically satisfied.
While these shortcut phenomena have different manifestations, both stem from models finding ways to satisfy constraints without learning intended semantic concepts.

This paper proposes a matrix-based differentiable logic programming method and 
presents systematic empirical analysis of how it addresses these shortcut phenomena. 
Building on recent work to encode logic programs as matrix operations \cite{takemuraECAI24}, we introduce technical refinements, establish connections to fuzzy logic semantics, and conduct experiments comparing against established NeSy baselines.

Prior work has explored the use of linear algebra to model logical inference in the context of logic programming.  
\cite{sakamaLinearAlgebraicCharacterization2017,sakamaLogicProgrammingTensor2021} introduced methods to represent propositional logic programs as matrix operations, enabling symbolic reasoning to be interpreted as fixed-point computations in vector spaces using numeric computation. 
\cite{aspisStableSupportedSemantics2020,takemuraGradientBasedSupportedModel2022,takemuraECAI24,satoEndtoEndASPComputation2026} proposed differentiable formulations of these matrix-based semantics, allowing integration with gradient-based learning. 
These methods support continuous-valued interpretations and enable symbolic logic to serve as a supervisory signal in neural architectures. 

The key insight underlying our approach is that matrix-based encoding establishes one-to-one correspondence between neural outputs and logical atoms, creating direct gradient paths from constraint violations to responsible neural predictions. 
This explicit connection contrasts with soft relaxation approaches where constraints may be satisfied through partial truth value assignments that do not commit to definitive concept classifications. 
We demonstrate through systematic experiments that this architectural choice significantly reduces both constraint satisfaction and cognition shortcuts compared to established NeSy baselines.

Our contributions are as follows:
\begin{itemize}
    \item We propose a matrix-based differentiable logic programming method to mitigate reasoning shortcuts, with systematic empirical analysis demonstrating that one-to-one atom grounding achieves better concept learning than fuzzy logic relaxations or probabilistic compilation.
    \item We introduce unified treatment of integrity constraints and implications via auxiliary atoms, reducing matrix storage while preserving supported model semantics.
    \item We identify connections between matrix operations and fuzzy logic t-norms and empirically compare how different aggregation strategies affect gradient properties in weakly supervised learning.
\end{itemize}

The rest of the paper is organized as follows. 
Section~\ref{sec:related-works} reviews related work, 
Section~\ref{sec:methodology} introduces the differentiable logic programming framework, 
Section~\ref{sec:shortcuts} describes the two shortcut phenomena, 
Section~\ref{sec:experiments} presents experimental results, 
and Section~\ref{sec:conclusion} presents the conclusions.

\section{Related Works}\label{sec:related-works}
Neurosymbolic AI has emerged as a paradigm for combining learning and reasoning \cite{hitzlerNeuroSymbolicArtificialIntelligence2022,hitzlerCompendiumNeurosymbolicArtificial2023}. 
Representative frameworks include DeepProbLog \cite{manhaeveDeepProbLogNeuralProbabilistic2018}, which couples neural training with probabilistic logic programming, DeepStochLog \cite{wintersDeepStochLogNeuralStochastic2022a}, which uses stochastic grammars for scalable inference, and NeurASP \cite{yangNeurASPEmbracingNeural2020}, which employs an Answer Set Programming (ASP) solver for gradient computation.

Differentiable reasoning approaches include Logic Tensor Networks (LTN) \cite{badreddineLogicTensorNetworks2022}, which integrates first-order logic with neural networks using fuzzy semantics, and Semantic Loss \cite{DBLP:conf/icml/XuZFLB18}, which relaxes logical operators via t-norm operations. 
These methods employ fuzzy logic semantics where truth values are continuous and logical operators are approximated by differentiable functions. 
In contrast, our matrix-based approach uses supported model semantics from logic programming \cite{satoEndtoEndASPComputation2026,takemuraECAI24}, where each matrix element corresponds to a specific ground atom with binary true/false assignments in the target model.

Our work builds on matrix-based logic programming semantics \cite{aspisStableSupportedSemantics2020,takemuraGradientBasedSupportedModel2022,takemuraECAI24}, which represent programs as matrix operations supporting differentiable fixed-point computations.
The advantage lies in explicit grounding: each matrix cell corresponds to a specific ground atom, creating direct gradient paths from constraint violations to neural predicates. 
In contrast, soft probability distributions or fuzzy logic relaxations \cite{badreddineLogicTensorNetworks2022,DBLP:conf/icml/XuZFLB18} can satisfy constraints through partial truth value assignments that do not require committing to definitive concept classifications, which may enable models to achieve high constraint satisfaction while learning incorrect concept mappings \cite{liLearningLogicalConstraints2022}.

Shortcut reasoning has recently gained attention in NeSy systems. 
\cite{liLearningLogicalConstraints2022} 
identified constraint satisfaction shortcuts and proposed regularization techniques, while our approach embeds logical semantics directly into optimization.
\cite{marconatoNotAllNeuroSymbolic2023} 
demonstrated that established NeSy systems are susceptible to reasoning shortcuts and proposed benchmark datasets \cite{marconatoBEARSMakeNeuroSymbolic2024a}.
Our contribution lies in providing systematic empirical analysis demonstrating 
how matrix-based semantics with explicit atom grounding addresses both shortcut types within a unified framework, whereas prior work on partial label learning \cite{wangLearningLatentModels2023}, continual learning \cite{marconatoNeuroSymbolicContinualLearning2023a}, and abductive learning \cite{yangAnalysisAbductiveLearning2024} each focused on shortcuts in specific learning settings with task-specific solutions.

\section{Differentiable Logic Programming}\label{sec:methodology}

In this section, we review the matrix-based differentiable logic programming framework from \cite{takemuraECAI24}, which forms the foundation of our approach. 
While the matrix encoding itself is established, our contribution lies in its application to shortcut mitigation and the design of systematic experiments demonstrating its effectiveness (Section~\ref{sec:experiments}).

\subsection{Logic Programs and Inference Semantics}

We consider \textit{propositional normal logic programs (NLPs)} $P$ consisting of rules of the form:
\begin{equation}\label{nlp}
    h \leftarrow b_1 \wedge \dots \wedge b_l \wedge \neg b_{l+1} \wedge \dots \wedge \neg b_m
\end{equation}
where $h$ and each $b_i$ are atoms from a Herbrand base $B_P$ of program $P$. 
The \textit{head} of a rule $r$ of the form~(\ref{nlp}) is denoted $head(r) = h$, and the \textit{body} is split into positive literals $body^+(r)=\{b_1,\ldots,b_l\}$ and negative literals $body^-(r)=\{b_{l+1},\ldots,b_m\}$.
A rule can have an empty head, and is called a \textit{constraint\/} in this case. 
An \textit{interpretation} $I \subseteq B_P$ \textit{satisfies} a rule $r$ if $body^+(r)\subseteq I$ and $body^-(r)\cap I=\emptyset$ together imply $head(r)\in I$. 
In particular, $I$ satisfies a constraint $r$ if it is not the case that $body^+(r)\subseteq I$ and $body^-(r)\cap I=\emptyset$, that is, if $body^+(r)\not\subseteq I$ or $body^-(r)\cap I\ne\emptyset$. 
An interpretation $I$ is a \textit{supported model} of a program $P$ if $I$ satisfies all rules of $P$ and for any atom $a\in I$, there is a rule $r\in P$ such that $head(r)=a$, $body^+(r)\subseteq I$ and $body^-(r)\cap I=\emptyset$. 

\subsection{Matrix Encoding of Logic Programs}\label{sec:subsec-matrix-encoding}
To enable differentiable evaluation of logic programs, we use a matrix encoding scheme based on prior work on linear algebraic representations of logic programs \cite{sakamaLogicProgrammingTensor2021,satoEndtoEndASPComputation2026,takemuraECAI24}.
While the matrix encoding follows prior works, our contribution lies in: (1) demonstrating how this encoding specifically addresses shortcut behaviors, (2) adapting the constraint handling for unified treatment of both shortcut types, and (3) empirical validation of its effectiveness compared to other NeSy approaches for shortcut mitigation.

Let $N$ be the number of atoms and $R$ the number of implication rules in a logic program $P$. 
We define:
\begin{itemize}
    \item Program matrix $\bm{Q} \in \{0,1\}^{R \times 2N}$: Each of the $R$ rows encodes one rule; the $2N$ columns correspond to the $N$ atoms and their $N$ negations, indicating which literals appear in the rule body.
    \item Head matrix $\bm{D} \in \{0,1\}^{N \times R}$: Each of the $N$ rows corresponds to an atom; each of the $R$ columns corresponds to a rule, with entry 1 indicating that the atom is the head of that rule.
\end{itemize}

We represent neural outputs as soft atom assignments using an \textit{interpretation vector} $\bm{v} \in [0,1]^N$ and its complement $\bm{w} = [\bm{v}; 1 - \bm{v}] \in [0,1]^{2N}$, covering both atoms and their negations.
Evaluation of the rule bodies is performed by computing \(\bm{h}\), which shows a soft derivation status of each atom, based on the current interpretation \(\bm{w}\).
\begin{equation}\label{eq:atom-satisfaction}
    \bm{h} = \min\left(1, \bm{D} \left(1 - \min\left(1, \bm{Q}(1 - \bm{w})\right)\right)\right)
\end{equation}
This equation evaluates rules in two steps: 
(1) $\bm{Q}(1 - \bm{w})$ computes body satisfaction for each rule by checking whether all required atoms hold and all negated atoms are false, and 
(2) $\bm{D}$ aggregates satisfied rule bodies to derive their corresponding head atoms.
The $\min$ operations ensure values remain in $[0,1]$.
An example is provided in Example \ref{ex:mnist-half}.

When encoding constraints, we use a semantically equivalent but syntactically different variant of the encoding in \cite{takemuraECAI24}.
Integrity constraints (IC), such as \(\leftarrow a \wedge b\), can be rewritten as \(z \leftarrow a \wedge b \wedge \neg z\) using an auxiliary atom \(z\) that must remain false.
This allows integrity constraints to be treated like any other implication rule in the matrix representation, unifying both in a single program matrix $\bm{Q}$. 
Prior work \cite{takemuraECAI24} required separate matrices for implications and constraints; our approach reduces storage from two matrices of sizes $(R \times 2N)$ and $(K \times 2N)$ to a single matrix of size $(R+K) \times (2N+2)$.

\subsection{Neural Atoms, Label Atoms and Loss Function}

Having defined the matrix encoding and constraint handling, we now describe how neural network outputs are integrated with the matrix evaluation to form a complete learning pipeline.

\begin{figure*}[t]
    \centering
    \includegraphics[width=0.9\linewidth]{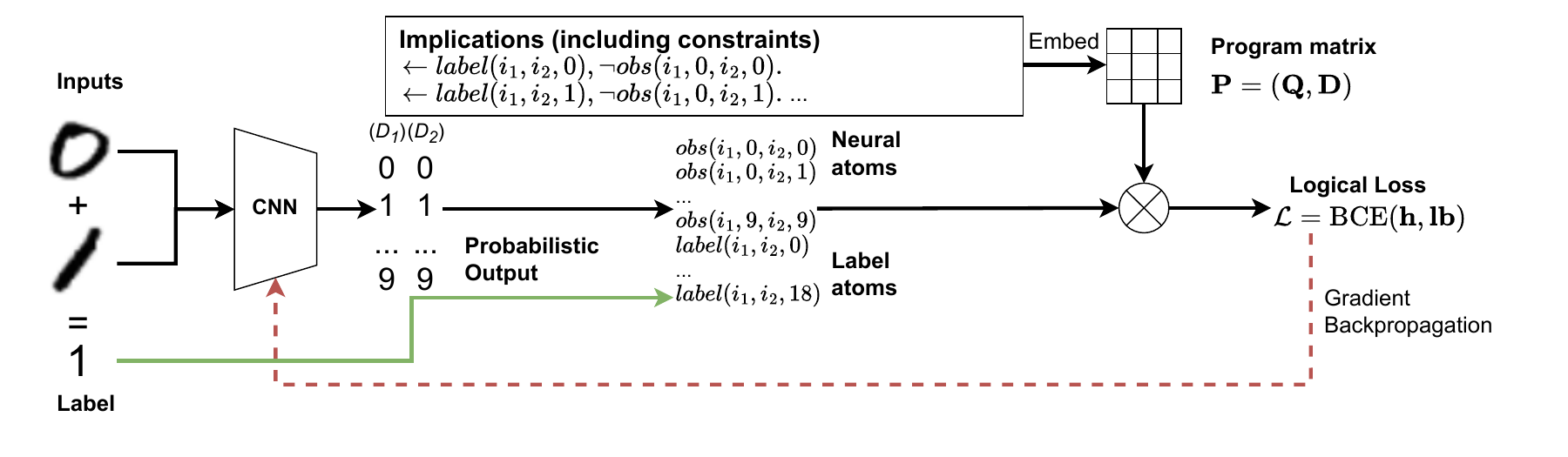}
    \caption{Learning pipeline for MNIST-Half.}
    \label{fig:pipeline}
\end{figure*}

Our learning pipeline for MNIST-Half \cite{marconatoBEARSMakeNeuroSymbolic2024a} is shown in Figure \ref{fig:pipeline}.
A neural network produces continuous truth values over atoms, forming the interpretation vector $\bm{v}$. 
These are mapped to the matrix-evaluation framework (Sec~\ref{sec:subsec-matrix-encoding}) using $\bm{w} = [\bm{v}; 1 - \bm{v}]$ (the extended vector containing both atoms and their negations) and are used to compute the soft derivation status of each atom, $\bm{h}$ (Eq \ref{eq:atom-satisfaction}).

The interpretation vector $\bm{h} \in [0,1]^N$ contains soft derivations for all $N$ atoms in the Herbrand base $B_P$ (Section~\ref{sec:subsec-matrix-encoding}), where each element $\bm{h}_i$ indicates how strongly atom $i$ is derived by the current rule evaluation.
In weakly supervised learning, the \textit{target vector} $\bm{t} \in \{0,1\}^M$ encodes supervision for $M$ atoms: one-hot encoded labels (e.g., sum labels 0-18) for label atoms, and $\bm{t}_z = 0$ for constraint auxiliary atoms to ensure constraint violations are penalized.
For example, in MNIST-Half \cite{marconatoBEARSMakeNeuroSymbolic2024a}, each training instance provides a \emph{ground truth sum} as the known result of adding two digit images (e.g., if the images depict digits 2 and 3, the ground truth sum is 5). 
The target vector $\bm{t}$ is a 20-dimensional vector: 19 elements encode the possible sum labels (0 to 18) as a one-hot vector, and 1 element sets the auxiliary atom $z$ to 0 (ensuring constraint violations are penalized).

We define the loss function as \(\mathcal{L} = \mathrm{BCE}(\bm{h}, \bm{t}) \), where $\mathrm{BCE}$ denotes the binary cross-entropy loss. 
This loss is minimized during training to encourage the neural network to output interpretations consistent with the logical rules.

\begin{example}[MNIST-Half]\label{ex:mnist-half}
    Consider MNIST-Half with 4 constraints (more details are available in Section \ref{sec:reasoning-shortcut}).
    This task resembles MNIST Addition, but with fewer and biased constraints that limit the range of observed combinations.
    We use two types of atoms to encode this task:
    \begin{itemize}
        \item $obs(i_1, d_1, i_2, d_2)$: True when images $i_1$ and $i_2$ are classified as digits $d_1$ and $d_2$ by the NN, respectively.
        \item $label(i_1, i_2, s)$: True when the ground truth sum for images $(i_1, i_2)$ is $s$.
    \end{itemize}
    
    The following 4 constraints encode the task:
    {\small
    \begin{align*}
        & (\textrm{R}1)\;\; \inlineimg{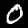} + \inlineimg{images/0.png} = 0  \;\;\;\; 
        (\textrm{R}2)\;\; \inlineimg{images/0.png} + \inlineimg{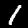} = 1 \\
        & (\textrm{R}3)\;\; \inlineimg{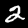} + \inlineimg{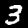} = 5 \;\;\;\;  
        (\textrm{R}4)\;\; \inlineimg{images/2.png} + \inlineimg{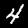} = 6 
    \end{align*}
    }
    In the propositional NLP form:
    {\small
    \begin{align*}
        &(\textrm{R}1)\;\;z \leftarrow label(i_1,i_2,0),\, \neg obs(i_1,0,i_2,0),\, \neg z. \; \\
        &(\textrm{R}2)\;\;z \leftarrow label(i_1,i_2,1),\, \neg obs(i_1,0,i_2,1),\, \neg z. \; \\
        &(\textrm{R}3)\;\;z \leftarrow label(i_1,i_2,5),\, \neg obs(i_1,2,i_2,3),\, \neg z. \; \\
        &(\textrm{R}4)\;\;z \leftarrow label(i_1,i_2,6),\, \neg obs(i_1,2,i_2,4),\, \neg z. \; \\
    \end{align*}
    }
    The program matrix $\bm{Q}$ (where $o$ stands for $obs$ atoms and $lb$ stands for $label$ atoms):
    {\scriptsize
    \begin{equation*}
    \let\quad\enspace
    \bordermatrix{ & o(i_1,0,i_2,0) & \cdots & lb(i_1,i_2,0) & \cdots & \neg o(i_1,0,i_2,0) & \cdots & \neg z \cr
                           (R1) & 0 & \cdots & 1 & \cdots      & 1        & \cdots  &  1  \cr
                           (R2) & 0 & \cdots & 0 & \cdots      & 0        & \cdots  &  1  \cr
                           (R3) & 0 & \cdots & 0 & \cdots      & 0        & \cdots  &  1  \cr   
                           (R4) & 0 & \cdots & 0 & \cdots      & 0        & \cdots  &  1  \cr   
    } \;
    \end{equation*}
    }
    Because all rules share the same head atom $z$, the head matrix $\bm{D}$ is a $(1 \times 4)$ row vector.
    {\scriptsize
    \begin{equation*}
    \bm{D} = \bordermatrix{ & (R1) & (R2) & (R3) & (R4) \cr
                            z & 1 & 1 & 1 & 1
    } \;
    \end{equation*}
    }
    This indicates that all four rules derive the same head atom $z$.

    Consider two cases for R1 (\inlineimg{images/0.png} + \inlineimg{images/0.png} = 0):

    (i) Misclassification (violation): 
    If the neural network misclassifies the first \inlineimg{images/0.png} as 1, then the atom $obs(i_1,0,i_2,0)$ becomes false (the actual network prediction is 1+0, not 0+0). 
    Given the ground truth $label(i_1,i_2,0)=1$ and $\neg obs(i_1,0,i_2,0)=1$ (negation of false), the rule body $\\ label(i_1,i_2,0) \wedge \neg obs(i_1,0,i_2,0) \wedge \neg z$ is satisfied, deriving $\bm{h}_z = 1$. 
    Since the target is $\bm{t}_z = 0$, the constraint is violated, yielding $\mathcal{L} = \mathrm{BCE}(1,0)$, which becomes numerically large.
    
    (ii) Correct classification (satisfied): 
    If the network correctly classifies both images as 0, then $obs(i_1,0,i_2,0)$ is true, making $\neg obs(i_1,0,i_2,0) = 0$. 
    This falsifies the rule body, yielding $\bm{h}_z = 0$, matching the target $\bm{t}_z = 0$, so $\mathcal{L} = \mathrm{BCE}(0,0) \approx 0$.
\end{example}

\subsection{Relationship to Fuzzy Logic Semantics}\label{sec:fuzzy-logic}

The matrix-based operations implicitly implement specific t-norm choices from fuzzy logic semantics, which affects gradient flow properties during training.

In fuzzy logic, conjunction is modeled via triangular norms (t-norms)  $\otimes: [0,1]^2 \to [0,1]$ \cite{DBLP:books/sp/KlementMP00}. 
Common t-norms include 
Product ($a \otimes b = ab$), 
G\"odel ($a \otimes b = \min(a,b)$), and 
{\L}ukasiewicz ($a \otimes b = \max(0, \allowbreak a+b-1)$).

The matrix encoding by \cite{takemuraECAI24} computes rule body satisfaction (a vector $\bm{sat} \in [0,1]^R$ indicating how well each rule's body is satisfied) via:
\begin{equation}
    \bm{sat} = 1 - \min(1, \bm{Q}(1 - \bm{w}))
\end{equation}
This operation corresponds to a \textit{bounded sum} aggregation over body literals where sums are capped at 1 to remain in $[0,1]$, which is equivalent to applying the {\L}ukasiewicz disjunction
followed by negation. 
For a conjunction $b_1 \wedge \cdots \wedge b_n$ with truth values $w_1, \ldots, w_n$ (elements of the extended interpretation vector $\bm{w}$ defined in Section \ref{sec:subsec-matrix-encoding}), we compute:
\begin{equation}
1 - \min\left(1, \sum_{i=1}^n (1-w_i)\right) = 
\max\left(0, \sum_{i=1}^n w_i - (n-1)\right)
\end{equation}
which is the {\L}ukasiewicz t-norm applied to all $w_i$.

Different t-norms exhibit distinct gradient flow properties: 
Product t-norm yields vanishing gradients when any $w_i \to 0$ (partially ameliorated by log-Product variants); 
G\"odel t-norm produces sparse gradients flowing only to the minimum element; 
{\L}ukasiewicz t-norm maintains non-zero gradients whenever $\sum w_i > (n-1)$.
We empirically compare these t-norm choices through comparison studies on MNIST Addition in Section \ref{sec:experiments}.

\section{Shortcut Behaviors in Neurosymbolic Systems}\label{sec:shortcuts}

Neurosymbolic systems are susceptible to two distinct forms of shortcut reasoning that undermine their intended learning objectives. 
While both phenomena result in models that satisfy logical constraints, they fail to acquire the semantic concepts those constraints were designed to enforce. 
In this section, we characterize these shortcut behaviors and provide illustrative tasks that demonstrate their occurrence.
In LP terms, these shortcuts correspond to situations where the trained neural network produces an interpretation that is a supported model of the program (satisfying all rules) but does not correspond to the \emph{intended} model, either because the interpretation trivially satisfies constraints (Section~\ref{sec:constraint-shortcut}) or because it realizes a different interpretation than the one the rules were designed to enforce (Section~\ref{sec:reasoning-shortcut}).

\subsection{Constraint Satisfaction Shortcuts}\label{sec:constraint-shortcut}

\subsubsection{Problem Characterization}

\textit{Constraint satisfaction shortcuts} \cite{liLearningLogicalConstraints2022} occur when models trivially satisfy logical constraints without learning intended semantic concepts. 
For a constraint $\leftarrow P\wedge\neg Q$ (equivalent to implication $P \rightarrow Q$, or $\neg P \vee Q$ in propositional logic), a model may learn to always output $\neg P$, satisfying the constraint while never learning concept $Q$. 
This is particularly problematic in semi-supervised settings where logical constraints provide the main supervision, leading to a mismatch between symbolic expectations and learned behavior.
In other words, the optimization finds a supported model that satisfies all integrity constraints but corresponds to an unintended interpretation. 

\subsubsection{Illustrative Task: MNIST 6}

To study this issue empirically, we adopt an MNIST classification task inspired by \cite{liLearningLogicalConstraints2022}. 
The goal is to train a digit classifier under a semi-supervised setting where one digit class is entirely unlabeled, and its learning is instead guided by a logical constraint.
In this setup, the labels of the class ``6'' are removed from all labeled training data. Instead, we provide the following logical constraint,

\begin{equation}\label{eq:missing6-cons1}
    (\neg \mathrm{NN}(R(x)) = 9 \; \vee \; \mathrm{NN}(x) = 6)
\end{equation}
where \(\mathrm{NN}\) refers to neural network classification, \(R(x)\) refers to the 180-degrees rotation of the input image \(x\).
This rule encodes the intuition that if a rotated image \(R(x)\) appears to be a ``9'' (\inlineimgrot{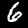}), then the original image $x$ is likely a ``6'' (\inlineimg{images/6.png}), or equivalently \(\mathrm{NN}(R(x))=9 \rightarrow \mathrm{NN}(x)=6\). 
However, the model may satisfy this constraint by simply learning to suppress the prediction ``9'' for all rotated images (i.e., making $\neg \mathrm{NN}(R(x))=9$ always true), which makes the implication vacuously true. 
In this case, the model never receives a gradient signal to learn what digit ``6'' looks like, so that the constraint is formally satisfied, but the intended semantic concept (the visual appearance of ``6'') is never acquired.
A complementary task where the missing digit is 9 (MNIST 9), is included in Appendix with experimental results.

\subsection{Cognition Shortcuts}\label{sec:reasoning-shortcut}

\subsubsection{Problem Characterization}

\textit{Cognition shortcuts} \cite{marconatoNotAllNeuroSymbolic2023} arise when neural components learn incorrect concept mappings despite formally satisfying logical rules. 
Unlike constraint satisfaction shortcuts, these occur even when rules are satisfied; the issue stems from biased training data causing the model to learn confounded mappings that satisfy rules syntactically but misalign semantically. 
We demonstrate this with MNIST-Half, with modulo addition shown in the Appendix.

\subsubsection{Illustrative Task: MNIST-Half}

We build on MNIST-Half from \cite{marconatoBEARSMakeNeuroSymbolic2024a}. 
In this task, the data distribution is biased such that only a subset of digits, specifically $\{0,1,2,3,4\}$, are used to create equations of the form \(\mathrm{NN}(x_1) + \mathrm{NN}(x_2) = \mathrm{label}\), where NN corresponds to neural network classification of images \(x_1\) and \(x_2\).
Unlike standard MNIST Addition \cite{manhaeveDeepProbLogNeuralProbabilistic2018}, which enumerates all digit pair combinations, MNIST-Half restricts training to a small subset of combinations, which deliberately introduces ambiguity.
More specifically, in the original configuration of MNIST-Half, only 4 constraints are allowed:
\begin{align*}
    (\textrm{R}1)\;\; \inlineimg{images/0.png} + \inlineimg{images/0.png} &= 0 \;\; &
    (\textrm{R}2)\;\; \inlineimg{images/0.png} + \inlineimg{images/1.png} &= 1 \\
    (\textrm{R}3)\;\; \inlineimg{images/2.png} + \inlineimg{images/3.png} &= 5 \;\; &
    (\textrm{R}4)\;\; \inlineimg{images/2.png} + \inlineimg{images/4.png} &= 6 
\end{align*}

Looking at the latter 2 rules (R3 and R4), even when confusing the concepts in the following manner, \(\{ \inlineimg{images/2.png} \mapsto 3, \inlineimg{images/3.png} \mapsto 2, \inlineimg{images/4.png} \mapsto 3 \}\), the model can satisfy the rules.
There are 3 possible mappings in this 4-rule scenario, out of which only 1 corresponds to the intended digit-to-label assignment; the other 2 are semantically incorrect yet still satisfy all 4 rules.
Thus, even though the rules (constraints) are seemingly satisfied during the training, the model's output after training may include incorrect mappings between the concepts and the intended labels.

\section{Experimental Evaluation}\label{sec:experiments}

In this section, we empirically evaluate our proposed method on two forms of shortcut reasoning in NeSy systems. 
The first subsection focuses on constraint satisfaction shortcuts (Section \ref{sec:constraint-shortcut}), and the second subsection focuses on cognition shortcuts (Section \ref{sec:reasoning-shortcut}). 
We compare our proposed method against representative neurosymbolic systems: Baseline CNN, Logic Tensor Networks (LTN) \cite{badreddineLogicTensorNetworks2022}, Semantic Loss \cite{DBLP:conf/icml/XuZFLB18}, Variational Learning \cite{liLearningLogicalConstraints2022}, DeepProbLog \cite{manhaeveDeepProbLogNeuralProbabilistic2018}, DeepStochLog \cite{wintersDeepStochLogNeuralStochastic2022a}, and NeurASP \cite{yangNeurASPEmbracingNeural2020}. 
Unless noted otherwise, all experiments use 10,000 training samples, Adam optimizer ($lr=10^{-3}$), 5 epochs, and average results over 10 runs.
We report digit classification accuracy as the primary metric because it directly measures concept learning quality, which is the core concern in shortcut mitigation.  
For constraint satisfaction shortcuts, we additionally report per-literal satisfaction rates (Tables~\ref{tab:missing-6} and~\ref{tab:missing-9}) to distinguish genuine concept learning from trivial constraint satisfaction.

\subsection{T-Norm Comparison}\label{sec:tnorm-comparison}

To empirically explore the relationship discussed in Section \ref{sec:fuzzy-logic}, we compare 4 alternative t-norm implementations ({\L}ukasiewicz, Product, Product-log, and G\"odel) on MNIST Addition.
Each t-norm evaluates conjunctions $b_1 \wedge \cdots \wedge b_n$ with truth values $w_1, \ldots, w_n$ as follows:
\begin{itemize}
    \item \textbf{{\L}ukasiewicz:} $\max(0, w_1 + \cdots + w_n - (n-1))$ (bounded sum)
    \item \textbf{Product:} $w_1 \times w_2 \times \cdots \times w_n$
    \item \textbf{Product-log:} $\exp(\sum_i \log w_i)$ (mathematically equivalent to Product, computed in log-space to avoid underflow)
    \item \textbf{G\"odel:} $\min(w_1, \ldots, w_n)$
    \item \textbf{MatLP \cite{takemuraECAI24}:} Matrix encoding (Eq.~\ref{eq:atom-satisfaction}), which uses bounded-sum aggregation equivalent to \\{\L}ukasiewicz with additional clamping operations.
\end{itemize}
The complete evaluation process proceeds in two steps: 
(1) apply the t-norm to evaluate each rule body $b_1 \wedge \cdots \wedge b_n$, producing body satisfaction values $\bm{sat} \in [0,1]^R$, then 
(2) for rules sharing the same head atom, aggregate their body satisfactions via the corresponding disjunction operator (maximum for G\"odel, bounded sum for {\L}ukasiewicz, probabilistic sum for Product) to compute the final head atom derivation $\bm{h}$.

Table~\ref{tab:fuzzy-mnist} shows that Product t-norm achieves the highest accuracy (91.7\%), followed by Product-log (90.5\%). 
The matrix-based method (88.5\%) and {\L}ukasiewicz (87.5\%) perform similarly, with the small difference likely due to clamping operations. 
G\"odel performs worst (82.9\%), confirming that sparse gradients (flowing only to the minimum element) hinder learning.
The strong performance of Product variants, despite theoretical gradient vanishing concerns, indicates that in tasks with rich supervision signals and short rule bodies, networks rarely encounter near-zero activations that would cause gradient issues. 
All t-norms show comparable training times ($\sim$155-171 seconds).

\begin{table}[htbp]
\centering
\caption{Fuzzy Operator Digit Accuracy on MNIST Addition}
\label{tab:fuzzy-mnist}
{\small
\begin{tabular}{@{\extracolsep{\fill}}l rr}
\toprule
Operator        & Acc. \%       & Time (s) \\
\midrule
G\"odel         & 82.9          & \underline{155.7} \\
{\L}ukasiewicz  & 87.5          & \textbf{155.0} \\
MatLP \cite{takemuraECAI24}  & 88.5          & 159.6 \\
Product         & \textbf{91.7} & 171.1 \\
Product-log     & \underline{90.5} & 156.1 \\
\bottomrule
\end{tabular}}
\end{table}

\subsection{Constraint Satisfaction Shortcuts}

We evaluate our method's ability to mitigate constraint satisfaction shortcuts using MNIST 6 and MNIST 9 described in Section~\ref{sec:constraint-shortcut}. 
Our goal is to determine whether the matrix-based framework encourages genuine concept learning rather than trivial constraint satisfaction.

\subsubsection{Experimental Setup}

We study constraint satisfaction shortcuts using a carefully designed semi-supervised classification task on MNIST \cite{lecunGradientbasedLearningApplied1998a} and USPS \cite{hullDatabaseHandwrittenText1994} datasets. 
Each training instance is a 4-tuple \((x_1, x_2, x_3, lb)\), where \(x_1\) is a labeled image, \(x_2\) is an unlabeled image that belongs to the missing class, \(x_3\) is a rotated version of the unlabeled image, and \(lb\) is the label of the labeled image. 
The missing-class digit (either ``6'' or ``9'') is never provided with a ground-truth label. 
Instead, its supervision is expected to be provided via a logical constraint involving the rotated counterpart (i.e., constraint (\ref{eq:missing6-cons1}) and its complement for MNIST 9)).

In addition to overall digit classification accuracy, we report the satisfaction rates of individual constraint literals on the test set for the missing digit class: 
(i) whether the rotated image avoids being classified as the counterpart digit (e.g., \(\neg \mathrm{NN}(R(x)) = 9\)),
and (ii) whether the unlabeled and unrotated image is correctly classified as the missing digit (e.g., 
\(\mathrm{NN}(x) = 6\)). 
These two metrics help distinguish between models that satisfy the constraint structurally and those that genuinely learn the intended concept.

\subsubsection{Results and Analysis}

The result for MNIST 6 is shown in Table \ref{tab:missing-6}. 
The baseline CNN model, which lacks logical supervision, consistently fails to learn the missing digit (0\% accuracy on the missing class), despite achieving moderate overall accuracy.
DeepProbLog, Semantic Loss, LTN and NeurASP satisfy the constraint structurally by suppressing the counterpart class in rotated images, but do not learn the target digit concepts. 

For example, Semantic Loss achieves $\neg \mathrm{NN}(R(x))=9$ satisfaction of 100\% (Table~\ref{tab:missing-6}) by distributing probability mass away from class 9 across all other classes, without concentrating it on any specific digit, which satisfies the constraint through partial truth values rather than committing to a definitive classification.
In contrast, our matrix-based encoding requires each atom to take a near-binary value, so the only way to satisfy the constraint is to actually classify the digit.

\begin{table}[htbp]
    \centering
    \caption{Digit Accuracy and Constraint Satisfaction Rates on MNIST 6.}
    \label{tab:missing-6}
    {\small
    \begin{tabular}{@{\extracolsep{\fill}}l rrr rrr}
    \toprule
     & \multicolumn{3}{c}{MNIST} &\multicolumn{3}{c}{USPS} \\
    \cmidrule(r){2-4} \cmidrule(l){5-7}
    Model & \footnotesize{Acc.\%} & \footnotesize{$\neg \mathrm{NN}(R(x))=9$} & \footnotesize{$\mathrm{NN}(x)=6$} & \footnotesize{Acc.\%} & \footnotesize{$\neg \mathrm{NN}(R(x))=9$} & \footnotesize{$\mathrm{NN}(x)=6$} \\
    \midrule
    Baseline CNN    & 88.4 & 33.0 & 0.0 & \underline{70.3} & 92.3 & 0.0 \\
    DPL             & 87.2 & 93.7 & 0.0 & 62.3 & \underline{99.1} & 0.0 \\
    DSL             & \textbf{97.0} & 35.0 & \textbf{97.7} & 59.5 & 92.4 & \textbf{56.6} \\
    LTN             & 60.5 & \underline{99.6} & 7.8 & 46.2 & \textbf{100.0} & 3.1 \\
    NeurASP         & 87.6 & 98.8 & 0.0 & 60.9 & 98.4 & 0.0 \\
    Semantic Loss    & 76.1 & \textbf{100.0} & 0.0 & 62.5 & \textbf{100.0} & 0.0 \\
    Variational L.  & 83.5 & 96.9 & 0.0 & 58.3 & 98.6 & 0.0 \\
    MatLP           & \underline{96.7} & 77.7 & \underline{92.3} & \textbf{71.5} & 97.5 & \underline{42.4} \\
    \bottomrule
    \end{tabular}}
\end{table}

DeepStochLog and our method achieve the highest classification accuracies on the missing digits, indicating successful concept learning guided by the constraint. 
Notably, our approach outperforms others in balancing constraint satisfaction with learning of intended concepts, where DeepStochLog and Variational Learning struggle to maintain generalization. 
These results highlight how existing NeSy systems can trivially satisfy constraints without learning the intended concepts, and demonstrate our method's ability to mitigate this shortcut behavior.
Thus, differentiable evaluation of logic rules can play a crucial role in reducing constraint satisfaction shortcuts of NeSy models.
Our matrix-based framework mitigates these issues by embedding logic programs in a differentiable form, allowing symbolic rules to influence the loss minimization process directly and encouraging models to satisfy constraints through intended semantics rather than trivial shortcuts.

\subsection{Cognition Shortcuts}

We now examine whether cognition shortcuts can be mitigated by progressively 
adding constraints that disambiguate confounded concepts, as described in Section~\ref{sec:reasoning-shortcut}. 
The hypothesis is that while Case 1 underdetermines concept mappings (allowing multiple semantically incorrect solutions that satisfy constraints), additional constraints in Cases 2-5 should progressively eliminate spurious mappings, enabling methods with explicit grounding to refine their concept learning.

\subsubsection{Experimental Setup}

We examine 5 variants of MNIST-Half, including the original setting, with increasing number of rules and label combinations to test whether cognition shortcuts can be mitigated through additional constraints:
\begin{enumerate}[label=Case~\arabic*:, leftmargin=*, align=left]
    \item Original setting: 4 rules (R1 to R4).
    \item To Case 1, add rule: $\inlineimg{images/0.png} + \inlineimg{images/2.png} = 2$
    \item To Case 2, add rule: $\inlineimg{images/0.png} + \inlineimg{images/3.png} = 3$
    \item To Case 2, add rule: $\inlineimg{images/0.png} + \inlineimg{images/4.png} = 4$
    \item To Case 2, add rules from Cases 3 and 4: $\inlineimg{images/0.png} + \inlineimg{images/3.png} = 3$ and $\inlineimg{images/0.png} + \inlineimg{images/4.png} = 4$
\end{enumerate}
The expectation is that, as we increase both the number of constraints and label information, that is, as we decrease the range of permissible combinations by the neural model, the resulting digit classification accuracy should improve.

We compare 7 models, including the baseline CNN and our proposed method, and all models have access to the identical constraint and label information.
We include two NeurASP variants: NeurASP uses fully grounded rules that enumerate all valid digit pairs explicitly (e.g., \texttt{addition(A,B,5) :- digit(0,A,2), digit(0,B,3).}), while NeurASP (org) uses the original compact encoding with arithmetic operators (e.g., \texttt{addition(A,B,N) :- digit(0,A,N1), digit(0,B,N2), N=N1+N2.}). 
The fully grounded variant makes each valid combination explicit in the program, analogous to our matrix-based encoding. 
The digit accuracy is calculated on the subset of MNIST, namely only using digits \(\{0,1,2,3,4\}\), as digits larger than 4 will not appear in the training set.

\subsubsection{Results and Analysis}

The results are shown in Table \ref{tab:reasoning-shortcut}.
Case 1 represents the original biased setting, where only a limited set of rules are provided during training.
Under this setup, most tested models, including LTN, DeepProbLog and DeepStochLog, fail to disambiguate the digit concepts.
NeurASP and our method perform better than those models, but despite better performance they are still affected by cognition shortcuts.
As additional constraints are introduced in Cases 2 to 5, performance diverges between models.
NeurASP shows a sharp improvement in Case 2, approaching perfect accuracy in Cases 3 through 5.
Our method demonstrates a more gradual but consistent improvement, reaching near perfect accuracy in Case 5.

These findings illustrate that symbolic constraints alone are not sufficient unless they cover the semantic ambiguities introduced by biased data.
When training distributions are biased, neural representations can collapse over semantically distinct concepts, resulting in logically consistent yet semantically incorrect inferences.
As additional constraints are introduced, methods with explicit atom grounding (NeurASP and our approach) show consistent improvement, while methods operating on continuous truth values through fuzzy logic relaxations (LTN, Semantic Loss) maintain high constraint satisfaction while learning incorrect concept mappings.

\begin{table}[htbp]
    \centering
    \caption{Test Accuracy (\%) on MNIST digits $\{0,1,2,3,4\}$ in MNIST-Half.}
    \label{tab:reasoning-shortcut}
    {\small
    \begin{tabular}{@{\extracolsep{\fill}}lrrrrr}
    \toprule
    Model & Case 1 & Case 2 & Case 3 & Case 4 & Case 5 \\
    \midrule
    Baseline & 21.1 & 26.2 & 35.4 & 35.9 & 30.4 \\
    DPL & 0.2 & 0.2 & 0.2 & 0.2 & 0.2 \\
    DSL & 0.4 & 0.4 & 0.9 & 1.0 & 1.0 \\
    LTN & 19.1 & 19.1 & 19.1 & 19.1 & 19.1 \\
    NeurASP & \underline{41.1} & \textbf{97.4} & \textbf{91.7} & \textbf{99.4} & 99.3 \\
    NeurASP (org) & 41.0 & 41.1 & \underline{91.5} & \underline{99.4} & \underline{99.5} \\
    Semantic Loss & 19.1 & 19.1 & 19.4 & 23.2 & 27.1 \\
    MatLP & \textbf{41.1} & \underline{68.3} & 80.1 & 81.8 & \textbf{99.5} \\
    \bottomrule
    \end{tabular}}
\end{table}

Table~\ref{tab:reasoning-shortcut} shows that several NeSy systems achieve surprisingly low accuracy on MNIST-Half Case 1. 
This is intentional: Case 1 severely underdetermines concept mappings (3 possible mappings for digits 2, 3, 4 with only 1 correct), whereas full MNIST Addition provides sufficient combinations to uniquely determine all digit mappings.

As we add constraints in Cases 2-5, we observe divergent behavior: DeepProbLog and DeepStochLog remain at low accuracy ($<2\%$), NeurASP successfully disambiguates concepts once sufficient constraints are provided (97\% in Case 2), and the matrix-based method shows gradual improvement (41\% to 99\%) demonstrating robust learning with incremental constraints. 
This pattern validates that MNIST-Half differentiates methods by shortcut resistance, where explicit grounding (NeurASP and the matrix-based approach) enables concept refinement with additional constraints, while soft relaxation methods struggle to escape initial shortcuts.

\subsubsection{Discussion}

MNIST-Half experiments reveal that methods with explicit atom grounding (NeurASP and the matrix-based approach) can leverage additional constraints to progressively refine concept mappings, while soft relaxation methods (LTN, Semantic Loss) and probabilistic compilation methods (DeepProbLog, DeepStochLog) struggle to escape initial shortcut solutions. 

However, explicit grounding alone does not guarantee shortcut mitigation.
In the constraint satisfaction experiments (Table~\ref{tab:missing-6}), NeurASP fails despite explicit grounding, while DeepStochLog succeeds via probabilistic exploration. 
This suggests that both representational structure (explicit atom grounding) and optimization mechanism (continuous differentiable evaluation vs.\ discrete enumeration) contribute to shortcut resistance.

\section{Conclusion}\label{sec:conclusion}

We proposed a matrix-based differentiable logic programming method to mitigate shortcut reasoning in NeSy systems and presented systematic empirical analysis of its effectiveness.
By encoding both implication rules and integrity constraints in a unified matrix representation, our approach establishes one-to-one correspondence between neural outputs and logical atoms.
Each matrix cell represents a specific ground atom, creating direct gradient paths from constraint violations to the responsible neural predictions.
This contrasts with fuzzy logic approaches where constraints are satisfied through partial truth values without definitive concept commitments.

Systematic experiments on both shortcut types demonstrate that this architectural choice significantly reduces shortcuts.
Methods with one-to-one atom grounding (our approach and NeurASP) progressively refine concept mappings as constraints are added, while fuzzy logic (LTN, Semantic Loss) and probabilistic compilation methods (DeepProbLog, DeepStochLog) struggle to escape initial shortcuts.
This reveals that coupling between symbolic knowledge and neural learning fundamentally affects shortcut mitigation: both explicit atom grounding and continuous differentiable optimization contribute to shortcut resistance, with neither alone being sufficient.

Our evaluation focuses on propositional programs over MNIST-based digit classification tasks. 
While this controlled setting effectively isolates shortcut behaviors for systematic analysis, the generalizability of our findings to more complex domains (e.g., real-world vision tasks \cite{marconatoNotAllNeuroSymbolic2023} or tasks requiring first-order reasoning) remains to be validated. 
Future work should address: scalability via sparse matrices, compositional reasoning benchmarks beyond vision, and extension to first-order logic with variables and quantifiers.

\section*{Acknowledgements}
This work has been supported by JSPS KAKENHI Grant Number JP25K03190 and JST CREST Grant Number JPMJCR22D3.

\bibliographystyle{eptcs}
\bibliography{ref}

\newpage
\appendix

\section{Additional Experiments}

\subsection{MNIST 6 Additional Baselines}

The following table shows the results for additional baselines which are omitted from Table \ref{tab:missing-6}.

\begin{table}[htbp]
    \centering
    \caption{Digit Accuracy and Constraint Satisfaction Rates on MNIST 6.}
    \label{tab:missing-6-baseline}
    {\small
    \begin{tabular}{@{\extracolsep{\fill}}l rrr rrr}
    \toprule
     & \multicolumn{3}{c}{MNIST} &\multicolumn{3}{c}{USPS} \\
    \cmidrule(r){2-4} \cmidrule(l){5-7}
    Model & \footnotesize{Acc.\%} & \footnotesize{$\neg \mathrm{NN}(R(x))=9$} & \footnotesize{$\mathrm{NN}(x)=6$} & \footnotesize{Acc.\%} & \footnotesize{$\neg \mathrm{NN}(R(x))=9$} & \footnotesize{$\mathrm{NN}(x)=6$} \\
    \midrule
    Baseline CNN    & 88.4 & 33.0 & 0.0 & 70.3 & 92.3 & 0.0 \\
    Baseline Linear & 11.9 & 93.3 & 17.8 & 12.8 & 77.2 & 24.5 \\
    Baseline MLP    & 13.0 & 73.0 & 7.0 & 12.6 & 75.0 & 7.8 \\
    Baseline SAN    & 14.3 & 77.8 & 11.8 & 13.4 & 79.1 & 8.6 \\
    \bottomrule
    \end{tabular}}
\end{table}

\subsection{MNIST 9} As a complementary test to MNIST 6, we reverse the setup: hide the digit ``9'' and provide a constraint:

\begin{equation}\label{eq:missing9}
    (\neg \mathrm{NN}(R(x)) = 6 \; \vee \; \mathrm{NN}(x) = 9)
\end{equation}
Here, the goal is to learn ``9'' (\inlineimg{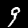}) via its relation to ``6'' (\inlineimgrot{images/9.png}), testing whether the model learns the intended concept or avoids learning to classify the digit ``9''.
The experimental results are shown in Table \ref{tab:missing-9}.

\begin{table}[htbp]
    \centering
    \caption{Digit Accuracy and Constraint Satisfaction Rates on MNIST 9.}
    \label{tab:missing-9}
    {\small
    \begin{tabular}{@{\extracolsep{\fill}}l rrr rrr}
    \toprule
     & \multicolumn{3}{c}{MNIST} &\multicolumn{3}{c}{USPS} \\
    \cmidrule(r){2-4} \cmidrule(l){5-7}
    Model & \footnotesize{Acc.\%} & \footnotesize{$\neg \mathrm{NN}(R(x))=6$} & \footnotesize{$\mathrm{NN}(x)=9$} & \footnotesize{Acc.\%} & \footnotesize{$\neg \mathrm{NN}(R(x))=6$} & \footnotesize{$\mathrm{NN}(x)=9$} \\
    \midrule
    Baseline CNN    & 88.1 & 18.1 & 0.0 & \textbf{75.9} & 26.0 & 0.0 \\
    Baseline Linear & 11.6 & 78.5 & 16.5 & 11.0 & 87.3 & 9.1 \\
    Baseline MLP    & 8.9 & 90.7 & 2.2 & 10.6 & 91.8 & 2.4 \\
    Baseline SAN    & 10.6 & 68.1 & 4.4 & 9.9 & 82.2 & 2.4 \\
    DPL             & 87.3 & 31.2 & 0.0 & 65.0 & 58.9 & 0.0 \\
    DSL             & \textbf{96.6} & 99.0 & \textbf{96.2} & 57.7 & \underline{99.9} & \underline{10.7} \\
    LTN             & 59.5 & \textbf{100.0} & 0.0 & 47.3 & \textbf{100.0} & 0.0 \\
    NeurASP         & 87.4 & \underline{99.3} & 0.0 & 63.5 & 92.9 & 0.0 \\
    Semantic Loss    & 77.2 & 99.0 & 0.0 & 62.0 & 93.3 & 0.0 \\
    Variational L.  & 83.4 & 98.3 & 0.0 & 62.2 & 60.4 & 0.0 \\
    MatLP           & \underline{93.7} & 97.9 & \underline{65.6} & \underline{72.2} & 87.1 & \textbf{19.4} \\
    \bottomrule
    \end{tabular}}
\end{table}

\subsection{Modulo Addition (mod k)}\label{sec:modulo-addition}

To further assess how NeSy systems generalize under varying complexity and limited supervision, we conduct experiments on the \textit{modulo addition task}.
Given a pair of digit images (\(x_1, x_2\)), the goal is to learn digit classification from the following equation:
\begin{equation}
    \mathrm{label} = (\mathrm{NN}(x_1)+\mathrm{NN}(x_2)) \; \mathrm{mod} \; k
\end{equation}
where NN denotes the neural network digit classification and \(k\) is a configurable value.
This task allows systematic variation in label ambiguity by changing \(k\) and training sample size, enabling fine-grained analysis of the effects of shortcut reasoning and generalization ability.

We consider values of \(k\) ranging from 2 to 10.
Lower values of \(k\) introduce more label collisions, increasing ambiguity; higher values of \(k\) increase the number of distinct output classes.

\begin{itemize}
    \item $k$: $2,...,10$
    \item Training data size: $100, 200, 400, 800, 1600, 3200, 6400, 12800, 25600$
    \item Number of runs: 10 for each $k$ and training data size combinations.
\end{itemize}

\subsubsection{Digit Classification Accuracy}

\begin{figure*}[h]
    \centering
    \includegraphics[width=0.95\linewidth]{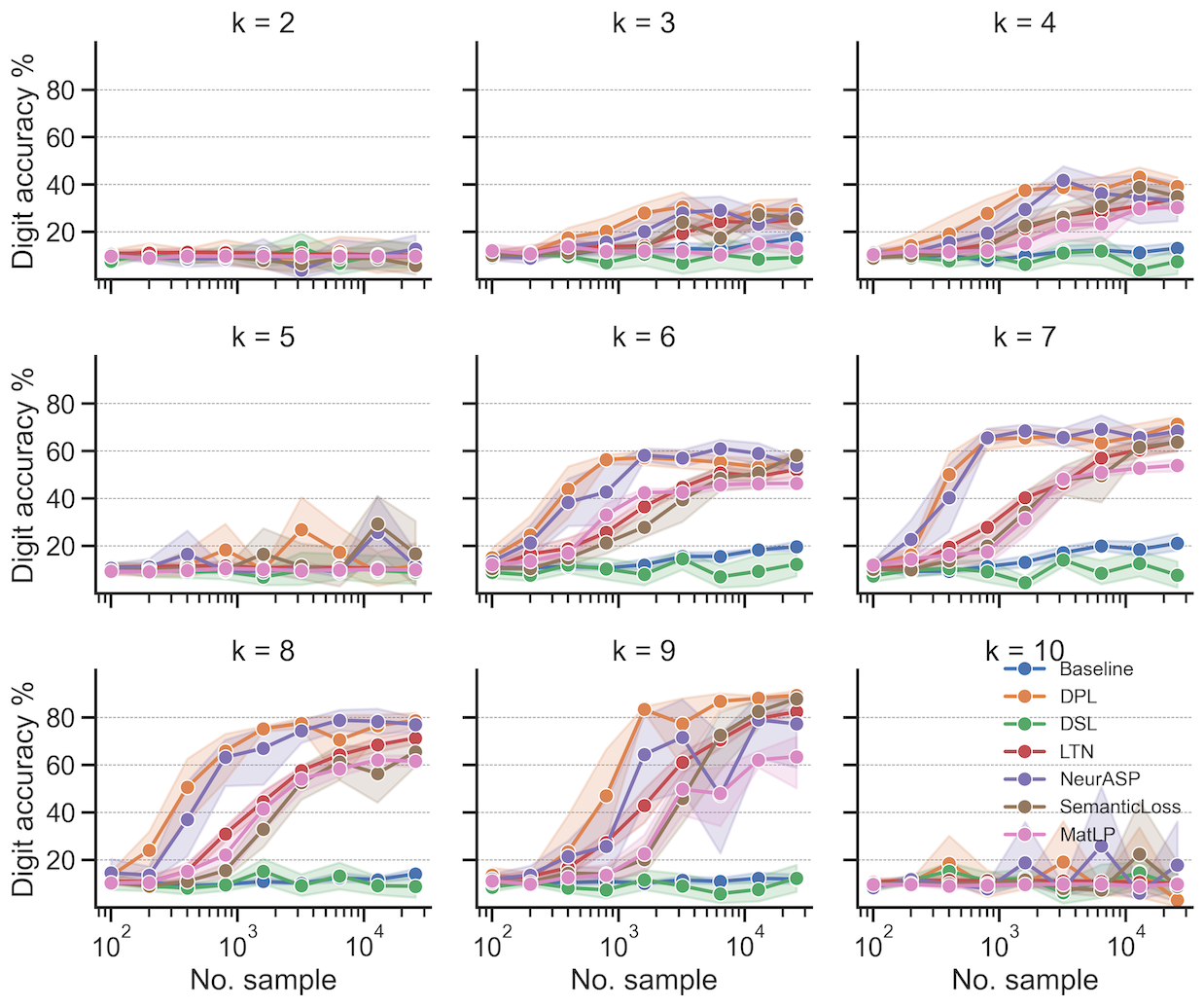}
    \caption{Digit classification accuracy in modulo addition task for varying mod~\(k\) and training sample size. DPL=DeepProbLog, DSL=DeepStochLog, NASP=NeurASP.}
    \label{fig:mod-k-accuracy}
\end{figure*}

Figure \ref{fig:mod-k-accuracy} shows digit classification accuracy across different values of mod \(k\) and training sample sizes.
This reflects how well each system learns the underlying digit concepts for correct addition.
As $k$ increases, the label space becomes more fine-grained and less ambiguous, and digit-level learning becomes more important. 
All methods except Baseline and DeepStochLog demonstrate improved digit classification as sample size increases, particularly for the intermediate values of $k$.

\subsubsection{Digit Classification Accuracy at Min/Max Training Data Sizes}

Tables \ref{tab:mod-k-accuracy-table-100} and \ref{tab:mod-k-accuracy-table-25600} show digit classification accuracy at the smallest and largest training sizes (100 and 25,600 examples), respectively, across $k=2$ to $k=10$.
This is intended to isolate the model performance at the smallest and full data availability.
For smaller values of $k$ (2–5), all models struggle, confirming that high label ambiguity leads to noisy supervision. 
For moderate values of $k$, DeepProbLog, Semantic Loss and NeurASP outperform others, likely due to their strong influence from the symbolic reasoners.
MatLP trails slightly, but maintains competitive results across the board.

\begin{table}[htbp]
\centering
\caption{Digit classification accuracy at Training Data Size = 100}
\label{tab:mod-k-accuracy-table-100}
{\small
\begin{tabular}{@{\extracolsep{\fill}}l rrr rrr rrr}
\toprule
Model / $K$ & 2 & 3 & 4 & 5 & 6 & 7 & 8 & 9 & 10 \\
\midrule
Baseline & 10.1 & 10.2 & \textbf{11.3} & 10.2 & 10.7 & 11.1 & 10.5 & 10.5 & \underline{10.3} \\
DeepProbLog & 9.8 & \underline{11.0} & 10.7 & 9.9 & \textbf{15.1} & \underline{11.8} & \underline{13.3} & \textbf{13.6} & 10.2 \\
DeepStochLog & 7.7 & 9.3 & 11.1 & \underline{11.0} & 8.8 & 7.4 & 10.6 & 8.5 & 8.8 \\
LTN & \textbf{10.9} & 10.2 & 9.7 & \textbf{11.1} & 10.8 & 10.0 & 11.0 & 11.7 & \textbf{10.6} \\
NeurASP & 9.8 & 10.0 & \underline{11.1} & 10.6 & \underline{13.5} & 11.6 & \textbf{14.5} & \underline{11.9} & 8.3 \\
Semantic Loss & \underline{10.5} & 10.1 & 9.1 & 9.4 & 10.5 & 10.1 & 10.3 & 10.4 & 10.0 \\
MatLP & 9.8 & \textbf{12.1} & 10.5 & 9.3 & 12.1 & \textbf{11.9} & 10.2 & 11.1 & 9.6 \\
\bottomrule
\end{tabular}}
\end{table}

\begin{table}[htbp]
\centering
\caption{Digit classification accuracy at Training Data Size = 25,600}
\label{tab:mod-k-accuracy-table-25600}
{\small
\begin{tabular}{@{\extracolsep{\fill}}l rrr rrr rrr}
\toprule
Model / $K$ & 2 & 3 & 4 & 5 & 6 & 7 & 8 & 9 & 10 \\
\midrule
Baseline & 9.8 & 17.4 & 13.2 & 9.7 & 19.6 & 21.1 & 14.1 & 11.9 & 9.8 \\
DeepProbLog & 9.2 & \textbf{29.2} & \textbf{39.2} & \underline{11.9} & \underline{55.8} & \textbf{71.3} & \textbf{78.8} & \textbf{89.5} & 3.1 \\
DeepStochLog & \underline{10.2} & 9.3 & 7.6 & 8.9 & 12.3 & 7.7 & 8.9 & 12.3 & 8.8 \\
LTN & 10.1 & 25.2 & 34.2 & 9.9 & 52.2 & 63.9 & 71.4 & 82.5 & \underline{10.1} \\
NeurASP & \textbf{12.8} & \underline{28.0} & 33.1 & 11.6 & 54.0 & \underline{68.4} & \underline{77.1} & 77.4 & \textbf{17.8} \\
Semantic Loss & 5.9 & 25.5 & \underline{35.0} & \textbf{16.7} & \textbf{58.2} & 63.8 & 65.6 & \underline{88.0} & 8.8 \\
MatLP & 9.8 & 13.0 & 30.4 & 10.0 & 46.4 & 54.0 & 61.6 & 63.5 & 9.8 \\
\bottomrule
\end{tabular}}
\end{table}

Tables \ref{tab:mod-k-time-table-100} and \ref{tab:mod-k-time-table-25600} shows the average training time across $k=2$ to $k=10$.
The training time in modulo addition do not depend on the values of $k$, and varies widely between different implementations.
While DeepProbLog can often outperform others in digit accuracy, it does so at the cost of being an order of magnitude slower than other models.

\begin{table}[htbp]
\centering
\caption{Training Time (seconds) at Training Data Size = 100}
\label{tab:mod-k-time-table-100}
{\small
\begin{tabular}{@{\extracolsep{\fill}}l rrr rrr rrr}
\toprule
Model / $K$ & 2 & 3 & 4 & 5 & 6 & 7 & 8 & 9 & 10 \\
\midrule
Baseline & \textbf{0.2} & \textbf{0.1} & \textbf{0.1} & \textbf{0.1} & \textbf{0.1} & \textbf{0.1} & \textbf{0.1} & \textbf{0.1} & \textbf{0.1} \\
DeepProbLog & 16.2 & 15.3 & 15.8 & 16.3 & 16.5 & 16.0 & 15.8 & 15.1 & 14.5 \\
DeepStochLog & 6.7 & 6.7 & 6.7 & 6.7 & 6.7 & 6.7 & 6.7 & 6.7 & 6.7 \\
LTN & 15.3 & 15.3 & 15.2 & 15.2 & 15.2 & 15.2 & 15.2 & 15.2 & 15.2 \\
NeurASP & 1.7 & 1.4 & 1.3 & 1.2 & 1.2 & 1.1 & 1.1 & 1.1 & 1.1 \\
Semantic Loss & 3.9 & 3.4 & 3.2 & 3.1 & 3.0 & 3.0 & 2.9 & 2.9 & 2.9 \\
MatLP & \underline{0.6} & \underline{0.6} & \underline{0.6} & \underline{0.6} & \underline{0.6} & \underline{0.6} & \underline{0.6} & \underline{0.6} & \underline{0.6} \\
\bottomrule
\end{tabular}}
\end{table}

\begin{table}[htbp]
\centering
\caption{Training Time (seconds) at Training Data Size = 25,600}
\label{tab:mod-k-time-table-25600}
{\small
\begin{tabular}{@{\extracolsep{\fill}}l rrr rrr rrr}
\toprule
Model / $K$ & 2 & 3 & 4 & 5 & 6 & 7 & 8 & 9 & 10 \\
\midrule
Baseline & \textbf{16.1} & \textbf{16.0} & \textbf{16.1} & \textbf{16.0} & \textbf{16.1} & \textbf{16.1} & \textbf{16.1} & \textbf{16.0} & \textbf{16.0} \\
DeepProbLog & 3789.7 & 3523.4 & 3444.5 & 4099.2 & 3405.0 & 3306.6 & 3222.9 & 3131.9 & 3439.2 \\
DeepStochLog & 408.1 & 408.5 & 408.4 & 407.6 & 407.8 & 411.8 & 407.8 & 407.2 & 409.9 \\
LTN & \underline{44.3} & \underline{44.3} & \underline{44.2} & \underline{44.2} & \underline{44.2} & \underline{44.2} & \underline{44.1} & \underline{44.1} & \underline{44.2} \\
NeurASP & 424.4 & 371.8 & 341.7 & 324.7 & 313.1 & 302.6 & 294.6 & 290.0 & 283.5 \\
Semantic Loss & 363.1 & 262.2 & 212.1 & 182.0 & 162.4 & 149.1 & 138.7 & 129.8 & 123.6 \\
MatLP & 143.3 & 143.9 & 144.2 & 142.5 & 144.2 & 144.1 & 143.9 & 143.3 & 144.1 \\
\bottomrule
\end{tabular}}
\end{table}

\subsubsection{Sum Accuracy}

\begin{figure*}[htbp]
    \centering
    \includegraphics[width=0.95\linewidth]{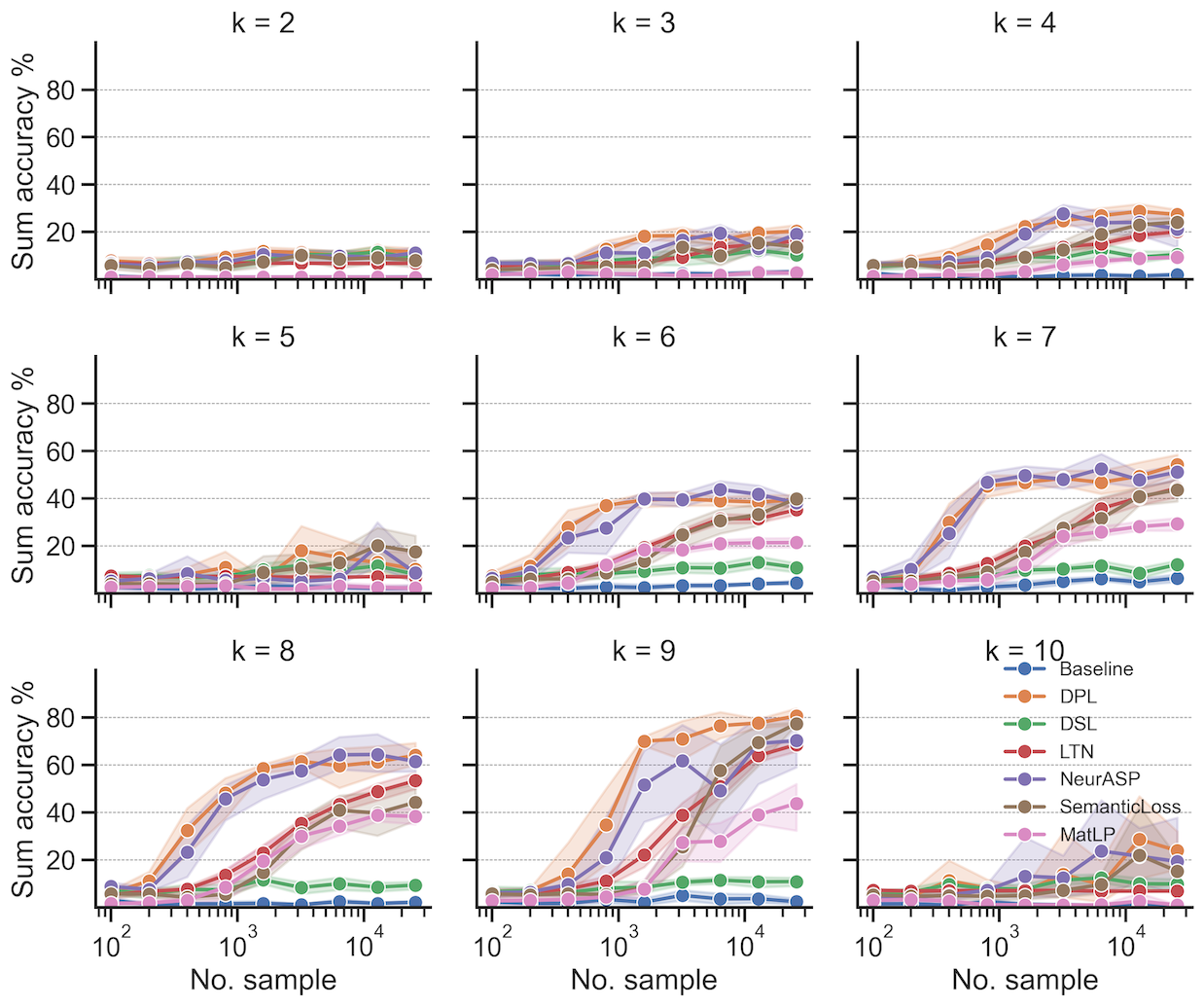}
    \caption{Sum accuracy \((\mathrm{NN}(x_1)+\mathrm{NN}(x_2))\) in modulo addition task.}
    \label{fig:mod-k-sum}
\end{figure*}

Figure \ref{fig:mod-k-sum} shows the accuracy of the unmodulated sum, \((\mathrm{NN}(x_1)+\mathrm{NN}(x_2))\), i.e., the model's ability to correctly learn addition before applying modulo.
A model may correctly predict the final label due to label coincidence, but still fail to capture the actual sum. 
Thus, high sum accuracy suggests better learning of intended digit concepts.
The overall trend is similar to digit accuracy, where all methods except Baseline CNN and DeepStochLog shows improved performance as the number of training example grows.



\subsection{Comparing Neural Baselines with `Logic' Components}

We conducted additional experiments on MNIST Addition comparing the matrix-based approach against neural alternatives (Linear, MLP, Outer product and Self-attention).
Out of these 4 alternatives, outer product most closely mimics how neural atoms are handled in the matrix-based method, and self-attention often results in competitive performance in supervised learning tasks.
The neural alternatives were not provided with information about the constraints in the form of logic programs, nevertheless, they all significantly underperformed the matrix-based approach.
This result further demonstrates incorporating external knowledge is necessary for effective learning in weakly supervised settings.

\begin{table}[htbp]
\centering
\caption{Digit Accuracy on MNIST Addition, Neural baselines}
\label{tab:neural-baseline-mnist}
{\small
\begin{tabular}{@{\extracolsep{\fill}}l rrr rrr rrr}
\toprule
Model & Acc. \% & Time (s) \\
\midrule
Linear          & \underline{13.2}  & 21.1 \\
MLP             & 9.0               & \underline{20.3} \\
Outer product   & 10.9              & \textbf{20.2} \\
Self-Attention  & 8.6               & 22.2 \\
\midrule
MatLP           & \textbf{95.4}     & 167.8 \\
\bottomrule
\end{tabular}}
\end{table}

\section{Neural Architectures}

\subsection{CNN Baseline}

All methods use the same base CNN architecture for fair comparison. 
We employ a LeNet style convolutional neural network for MNIST digit classification:

\begin{itemize}
    \item Input: $28 \times 28$ greyscale images
    \item Conv2D(6, 5)
    \item MaxPool2D(2, 2)
    \item ReLU
    \item Conv2D(16, 5)
    \item MaxPool2D(2, 2)
    \item ReLU
    \item Linear(120)
    \item ReLU
    \item Linear(84)
    \item ReLU
    \item Linear(10)
    \item Output: Softmax
\end{itemize}

The dimension of the final layer may change when applied to variants of MNIST Addition, e.g., for a single digit addition, the final layer is Linear(19) for directly predicting the sum.

\subsection{Other Baseline Architectures}

\paragraph{Linear}
Used in MNIST 6, MNIST 9 and neural-baseline comparison (MNIST Addition) experiments. 
The network consists of the CNN architecture followed by a Linear(10) layer for reasoning.

\paragraph{MLP}
Used in MNIST 6, MNIST 9 and neural-baseline comparison (MNIST Addition) experiments.
The network consists of the CNN architecture followed by an MLP (multilayer perceptron) for reasoning.
MLP consists of Linear(32), ReLU, Linear(16), ReLU and Linear(10) layers.

\paragraph{SAN}
Used in MNIST 6, MNIST 9 and neural-baseline comparison (MNIST Addition) experiments.
The network consists of the CNN architecture followed by a self-attention layer for reasoning.
The self-attention layer is a MultiheadAttention layer with 2 heads and 10-dim dimension.
The heads consist of Linear(32), ReLU and Linear(10).

\paragraph{Outer product}
Used in neural-baseline comparison experiment (MNIST Addition).
The network consists of the CNN architecture followed by a layer which computes the outer product followed by a Linear layer.

\section{Task-Specific Encodings}

This section documents the encodings used in each method for comparison.
Note that these are only \textit{partial} encodings highlighting the essential parts related to each task.

\subsection{MNIST 6}

Each training instance contains: 
(1) labeled image $x_1$ from classes $\{0,1,2,3,4,5,7,8,9\}$ (excluding the missing class), 
(2) unlabeled image $x_2$ from the missing class (6), and 
(3) rotated image $x_3 = R_{180}(x_2)$. 
The constraint enforces that if $R(x)$ is classified as the complement digit, then $x$ should be classified as the missing digit.
The missing class is 6 and 9 for MNIST 6 and MNIST 9, respectively.

\textbf{DeepProbLog}

\begin{lstlisting}[language=Prolog, basicstyle=\footnotesize\ttfamily]
rotate(X,Z) :- rdigit(Z,0); rdigit(Z,1); rdigit(Z,2); rdigit(Z,3); 
               rdigit(Z,4); rdigit(Z,5); rdigit(Z,6); rdigit(Z,7); 
               rdigit(Z,8); digit(X,6).
               
combine(X,Z,Y) :- rotate(X,Z), digit(X,Y).
\end{lstlisting}

\textbf{DeepStochLog}

\begin{lstlisting}[language=Prolog, basicstyle=\footnotesize\ttfamily]
rotate(6) --> rdigit(9).
rotate(Z) --> rdigit(Z), { Z \= 9 }.

combine(X) --> digit(Y), rotate(Z), { X = Y }.
\end{lstlisting}

\textbf{Logic Tensor Networks}

\texttt{image\_i1} denotes the labeled image. 
\texttt{image\_r1} and \texttt{image\_u1} refer to rotated and unlabaled images, respectively.

\begin{lstlisting}[language=Python, basicstyle=\footnotesize\ttfamily]
sat_agg = SatAgg(
    Forall(image_i1, D(image_i1, label_l1)),
    Forall(ltn.diag(image_r1, image_u1),
           Implies(D(image_r1, c_9), D(image_u1, c_6))))
\end{lstlisting}

\textbf{NeurASP}

\texttt{rdigit} and \texttt{udigit} refer to rotated and unlabaled digits, respectively.

\begin{lstlisting}[language=Prolog, basicstyle=\footnotesize\ttfamily]
:- rdigit(0,B,9), not udigit(0,B,6).
\end{lstlisting}

\textbf{Semantic Loss}

\begin{lstlisting}[language=Python, basicstyle=\footnotesize\ttfamily]
loss2 = (- torch.log(probs_u[:,6] + 1e-8) 
         - torch.log(1-probs_u[:, :] + 1e-8).sum(dim=-1) 
         + torch.log(1-probs_u[:,6] + 1e-8))
         
loss3 = (- torch.log(probs_r[:,9] + 1e-8) 
         - torch.log(1-probs_r[:, :] + 1e-8).sum(dim=-1)
         + torch.log(1-probs_r[:,9] + 1e-8))

\end{lstlisting}

\textbf{Matrix-based}

The rotation rule below is for illustration purposes only. The actual encoding is in the matrix format.
\begin{lstlisting}[language=Python, basicstyle=\footnotesize\ttfamily]
z :- rotated(img2, 9), not normal(img1, 6), not z.
\end{lstlisting}

\subsection{MNIST-Half}

Each training instance contains two images $(x_1, x_2)$ and sum label $s$. 
The constraint enforces $\\ \texttt{digit}(x_1, d_1) + \texttt{digit}(x_2, d_2) = s$.

\textbf{DeepProbLog}

\begin{lstlisting}[language=Prolog, basicstyle=\footnotesize\ttfamily]
addition(X,Y,Z) :- digit(X,X2), digit(Y,Y2), Z is X2+Y2.
\end{lstlisting}

\textbf{DeepStochLog}

\begin{lstlisting}[language=Prolog, basicstyle=\footnotesize\ttfamily]
addition(N) --> is_number(N1), is_number(N2), {N is N1 + N2}.
\end{lstlisting}

\textbf{Logic Tensor Networks}

\begin{lstlisting}[language=Python, basicstyle=\footnotesize\ttfamily]
sat_agg = Forall(
    ltn.diag(images_x, images_y, labels_z),
    Exists(
        [d_1, d_2],
        And(Digit_s_d(images_x, d_1), Digit_s_d(images_y, d_2)),
        cond_vars=[d_1, d_2, labels_z],
        cond_fn=lambda d1, d2, z: torch.eq(d1 + d2, z),
        p=p
    ))
\end{lstlisting}

\textbf{NeurASP}

\begin{lstlisting}[language=Prolog, basicstyle=\footnotesize\ttfamily]
addition(A,B,0) :- digit(0,A,0), digit(0,B,0).

addition(A,B,1) :- digit(0,A,0), digit(0,B,1).

addition(A,B,5) :- digit(0,A,2), digit(0,B,3).
addition(A,B,5) :- digit(0,A,3), digit(0,B,2).
addition(A,B,5) :- digit(0,A,4), digit(0,B,1).

addition(A,B,6) :- digit(0,A,2), digit(0,B,4).
addition(A,B,6) :- digit(0,A,3), digit(0,B,3).
addition(A,B,6) :- digit(0,A,4), digit(0,B,2).
\end{lstlisting}

\textbf{NeurASP original encoding}

\begin{lstlisting}[language=Prolog, basicstyle=\footnotesize\ttfamily]
addition(A,B,N) :- digit(0,A,N1), digit(0,B,N2), N=N1+N2.
\end{lstlisting}

\textbf{Semantic Loss}

\begin{lstlisting}[language=Python, basicstyle=\footnotesize\ttfamily]
for d1 in valid_digits:
    for d2 in valid_digits:
        if d1 + d2 == target_sum:
            valid_pair_prob += probs_d1[b, d1] * probs_d2[b, d2]

# Negative log probability
constraint_loss[b] = -torch.log(valid_pair_prob + 1e-8)
\end{lstlisting}

\textbf{MatLP}

The rules below are for illustration purposes only. The actual encoding is in the matrix format.
\begin{lstlisting}[language=Python, basicstyle=\footnotesize\ttfamily]
z :- label(img1, img2, 0), not obs(img1, 0, img2, 0), not z.
z :- label(img1, img2, 1), not obs(img1, 0, img2, 1), not z.
z :- label(img1, img2, 5), not obs(img1, 2, img2, 3), not z.
z :- label(img1, img2, 6), not obs(img1, 2, img2, 4), not z.
\end{lstlisting}

\subsection{Mod K Addition}

Given a pair of digit images (\(x_1, x_2\)), the goal is to learn digit classification from the following equation:
\begin{equation}
    \mathrm{label} = (\mathrm{NN}(x_1)+\mathrm{NN}(x_2)) \; \mathrm{mod} \; k
\end{equation}

\textbf{DeepProbLog}

\begin{lstlisting}[language=Prolog, basicstyle=\footnotesize\ttfamily]
addition(X,Y,K,Z) :- digit(X,X2), digit(Y,Y2), Z is (X2+Y2) mod K.
\end{lstlisting}

\textbf{DeepStochLog}

\begin{lstlisting}[language=Prolog, basicstyle=\footnotesize\ttfamily]
addition(N, K) --> is_number(N1), is_number(N2), {N is (N1 + N2) mod K}.
\end{lstlisting}

\textbf{Logic Tensor Networks}

\begin{lstlisting}[language=Python, basicstyle=\footnotesize\ttfamily]
sat_agg = Forall(
    ltn.diag(images_x, images_y, labels_z),
    Exists(
        [d_1, d_2],
        And(Digit_s_d(images_x, d_1), Digit_s_d(images_y, d_2)),
        cond_vars=[d_1, d_2, labels_z],
        cond_fn=lambda d1, d2, z: torch.eq((d1 + d2) % K, z),
        p=p
    ))
\end{lstlisting}

\textbf{NeurASP}

\begin{lstlisting}[language=Prolog, basicstyle=\footnotesize\ttfamily]
addition(A,B,N) :- digit(0,A,N1), digit(0,B,N2), N=(N1+N2) \ K.
\end{lstlisting}

\textbf{Semantic Loss}

\begin{lstlisting}[language=Python, basicstyle=\footnotesize\ttfamily]
for d1 in range(10):
    for d2 in range(10):
        if (d1 + d2) % self.k == target_mod:
            valid_pair_prob += probs_d1[b, d1] * probs_d2[b, d2]
\end{lstlisting}

\textbf{MatLP}

The addition rule below is for illustration purposes only. The actual encoding is in the matrix format.
\begin{lstlisting}[language=Python, basicstyle=\footnotesize\ttfamily]
z :- label(img1, img2, 0 \ K), not obs(img1, 0, img2, 0), not z.
\end{lstlisting}

\section{Experimental Details}

\subsection{Configurations and Hyperparameters}

\paragraph{Common Settings (All Experiments).}
\begin{itemize}
    \item \textbf{Optimizer}: Adam \cite{kingma2014adam}
    \item \textbf{Learning rate}: $10^{-3}$ (fixed, no scheduler)
    \item \textbf{Batch size}: 32 (where available)
    \item \textbf{Training data}: 10,000 samples
    \item \textbf{Epochs}: 5
    \item \textbf{Number of runs}: 10
    \item \textbf{Random seeds}: ${2025, 2026, 2027, ..., 2034}$ (seed for each run)
    \item \textbf{Timeout}: None
\end{itemize}

\subsection{Computing Environment}

\paragraph{Common Environment (All Experiments).}
\begin{itemize}
    \item \textbf{CPU}: AMD Ryzen 9 7950X (16 cores)
    \item \textbf{RAM}: 128GB
    \item \textbf{GPU}: NVIDIA RTX A4000 16GB
    \item \textbf{OS}: Ubuntu 22.04 LTS
    \item \textbf{Python}: 3.10.11
    \item \textbf{PyTorch}: 2.0.1 with CUDA 11.7
\end{itemize}

\subsection{Code Availability}

Codes used in experiments are based on the following repositories:

\begin{itemize}
    \item \textbf{DeepProbLog}: \url{https://github.com/ML-KULeuven/deepproblog}
    \item \textbf{DeepStochLog}: \url{https://github.com/ML-KULeuven/deepstochlog}
    \item \textbf{Logic Tensor Networks}: \url{https://github.com/tommasocarraro/LTNtorch}
    \item \textbf{NeurASP}: \url{https://github.com/azreasoners/NeurASP}
    \item \textbf{Semantic Loss}: \url{https://github.com/UCLA-StarAI/Semantic-Loss}
    \item \textbf{Semantic Loss (Variational Loss)}: \\ \url{https://github.com/SoftWiser-group/NeSy-without-Shortcuts}
    \item \textbf{MatLP}: \url{https://github.com/atakemura/difflogic}
\end{itemize}

\end{document}